\documentclass[%
 reprint,
 jou,
]{apa6}%{revtex4-2}

\pdfoutput=1
\usepackage{apacite}
\usepackage{amsmath}
\usepackage{amssymb}
\usepackage[normalem]{ulem}

\usepackage{url}
\usepackage{breakurl}

\usepackage{float}
\usepackage{blindtext}
\usepackage[utf8]{inputenc}
\usepackage{graphicx}% Include figure files
\usepackage{dcolumn}% Align table columns on decimal point
\usepackage{bm}% bold math

\usepackage[dvipsnames]{xcolor}
\title{Scale-Dependent Relationships in Natural Language}% Force line breaks with \\
\shorttitle{Scale-Dependent Relationships in Natural Language}

\author{Aakash Sarkar, Marc Howard}
\affiliation{%
 Department of Psychological and Brain Sciences\\
 Boston University
}%

\date{\today}% It is always \today, today,
\begin{document}
\maketitle
\begin{abstract}

Natural language exhibits statistical dependencies at a wide range of scales.  
For instance, the mutual information between words in natural language decays like a power law with the temporal lag between them.
However, many statistical learning models applied to language impose a sampling scale while extracting statistical structure. 
For instance, Word2Vec constructs a vector embedding that maximizes the prediction between a target word and the context words that appear nearby in the corpus.
The size of the context is chosen by the user and defines a strong scale; relationships over much larger temporal scales would be invisible to the algorithm.
This paper examines the family  of Word2Vec embeddings generated while systematically manipulating the sampling scale used to define the context around each word. 
The primary result is that different linguistic relationships are preferentially encoded at different scales. 
Different scales emphasize different syntactic and semantic relations between words.
Moreover, the neighborhoods of a given word in the embeddings change significantly depending on the scale. 

These results suggest that any individual scale can only identify a subset of the meaningful relationships a word might have, and point toward the importance of developing scale-free models of semantic meaning.
\end{abstract}

\section{Introduction}

Information in natural sequences often spans across many scales. A mixture of many length scales have been seen to create a  power-law decay of long-range correlations in DNA sequences \cite{LiEtal94,PengEtal92,MantEtal94}. Compositions from different composers in Western classical music obey a $1/f^{\alpha}$ power law in both musical pitch and rhythm spectra \cite{LeviEtal12, RoosMana07}. Such scale-free behavior has been observed in earthquakes \cite{AbeSuzu05}, collective motion of starling flocks \cite{CavaEtal10a}, and neural amplitude fluctuations in the human brain \cite{LinkEtal01}. Samples of natural language also exhibit long-range fractal correlations \cite{MontPury02}. The mutual information (MI) between two symbols, for such sequences, have recently been shown to decay like a power law as well, with the temporal difference between them \cite{LinTegm16} (see Figure 1) .

Analyses on large-text corpora from diverse sources have been shown to have long-range structure beyond the short-range correlations happening at syntactic level between sentences \cite{EbelNeim95,EbelPosc94}. Corpora from different languages have been shown to have a two-scale structure, with the dimension of semantic spaces at short distances being distinctly smaller than at long distances \cite{DoxaEtal10}. Studies on the statistics of shuffled text corpora seem to confirm this, where a text corpora shuffled even at the sentence level loses large-scale structure \cite{AltmEtal12}. However, many prevalent statistical learning models which aim to learn such semantic structure fix a scale when sampling the context around words. We observe one such class of models called Word2Vec, which use a vector embedding to study semantic structure. Word2Vec uses a moving window around each word to gather context, but the size of the window is a fixed parameter. In this paper we explore how changing this size of sampling context can change the structure of the embedding, and what it means for the information it encodes about the training text.

\begin{figure*}[!ptb]
\begin{center}
\minipage{0.85\linewidth}
    
    \includegraphics[width=0.99\columnwidth]{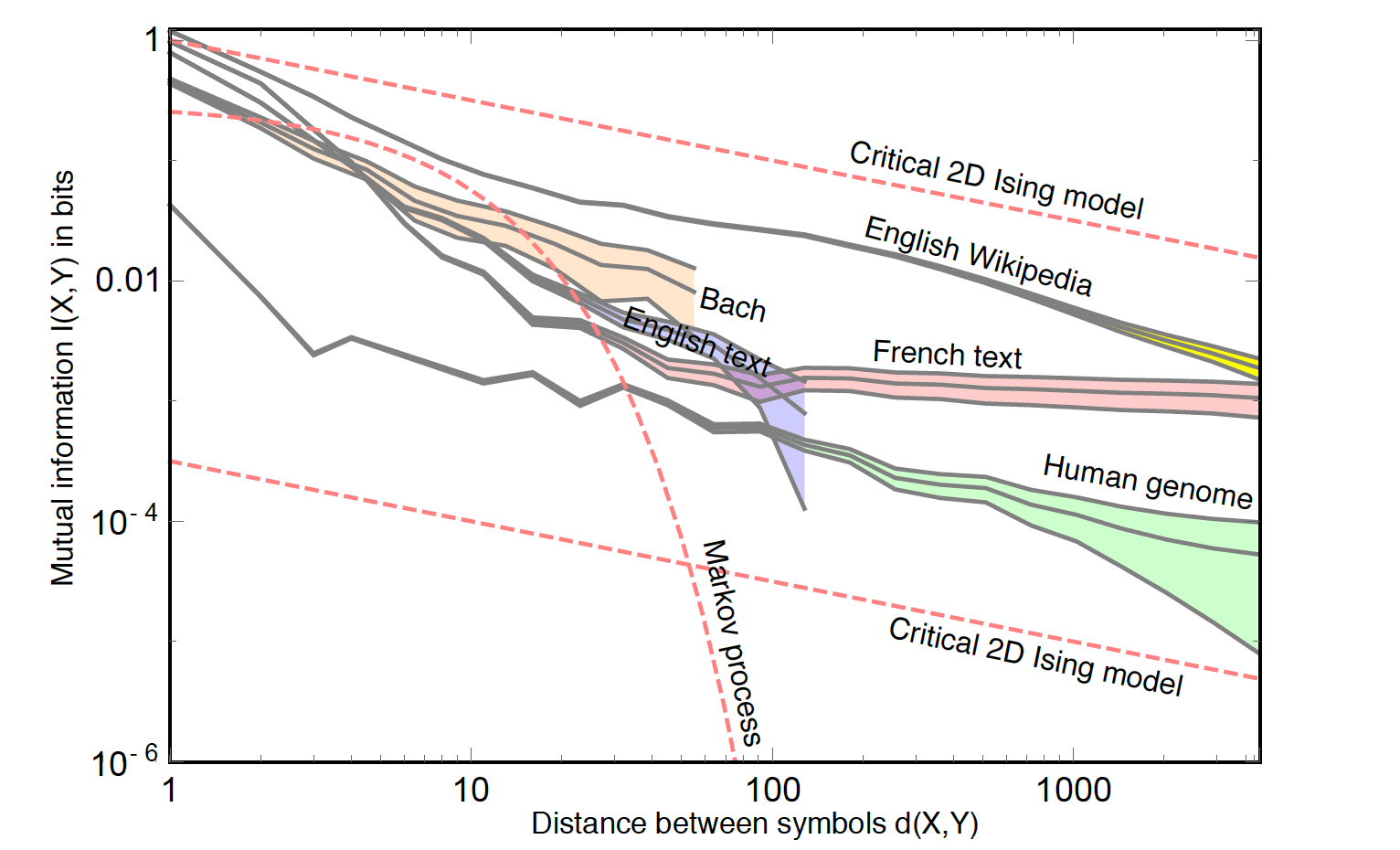}    % 
    %  \scriptsize{From McCormick, C., 2016, Word2Vec Tutorial}   

\endminipage
\end{center}
\caption{
\emph{Language has information at many scales.} Mutual information (MI) between a pair of symbols in different natural sequences falls slowly as a function of how far they are spaced \protect\cite{LinTegm16}. The MI is a measure of the shared information content between the two symbols, and this seems to decay roughly as a power law for natural language. This is contrasted with the sharp exponential fall seen by a Markov process which has a fixed, predetermined scale. The slow decay of MI suggests that information is contained at a spectrum of different scales, and algorithms sampling natural language at fixed scales might not be sufficient.}
\end{figure*}
\label{fig:mutual}

\subsection{Word2Vec and Vector Embeddings}

Word2Vec is a widely used neural network model which learns a vector representation of words, called an embedding, by training on large corpora of text. Word embeddings store a unique vector representation of each word in the vocabulary in a high-dimensional vector space - a good embedding would map semantically similar words onto nearby points onto this vector space. Analyzing the structure of the embedding should also provide insight into the relations between words and how they appear in the source corpus.

Word2Vec is a predictive model which tries to infer a relationship between a central word, referred to as \emph{target}, and its surrounding words, referred to as \emph{context}. It comes in two flavors, which use the same algorithm but act as inverses of each other. The Skip-gram model tries to predict the context words from the target word, and the Continuous Bag-of-Words (CBOW) model tries to predict the target word from the context words around it. In both cases, the training continuously modifies the embedding with each target and context set, so that it would maximize the probability of obtaining one from the other (depending on the flavor). In this article, we focus on the CBOW variant and the structure of the embeddings it generates.

\begin{figure*}[!htb]
\begin{center}
\minipage{0.475\linewidth}
\begin{center}
    \includegraphics[width=0.99\columnwidth]{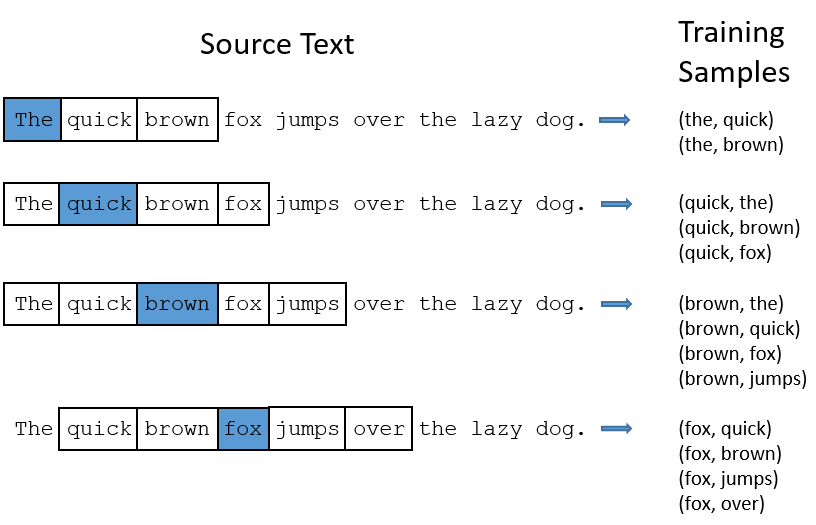}

\end{center}
\endminipage
\minipage{0.47\linewidth}
\begin{center}
    \includegraphics[width=0.95\textwidth]{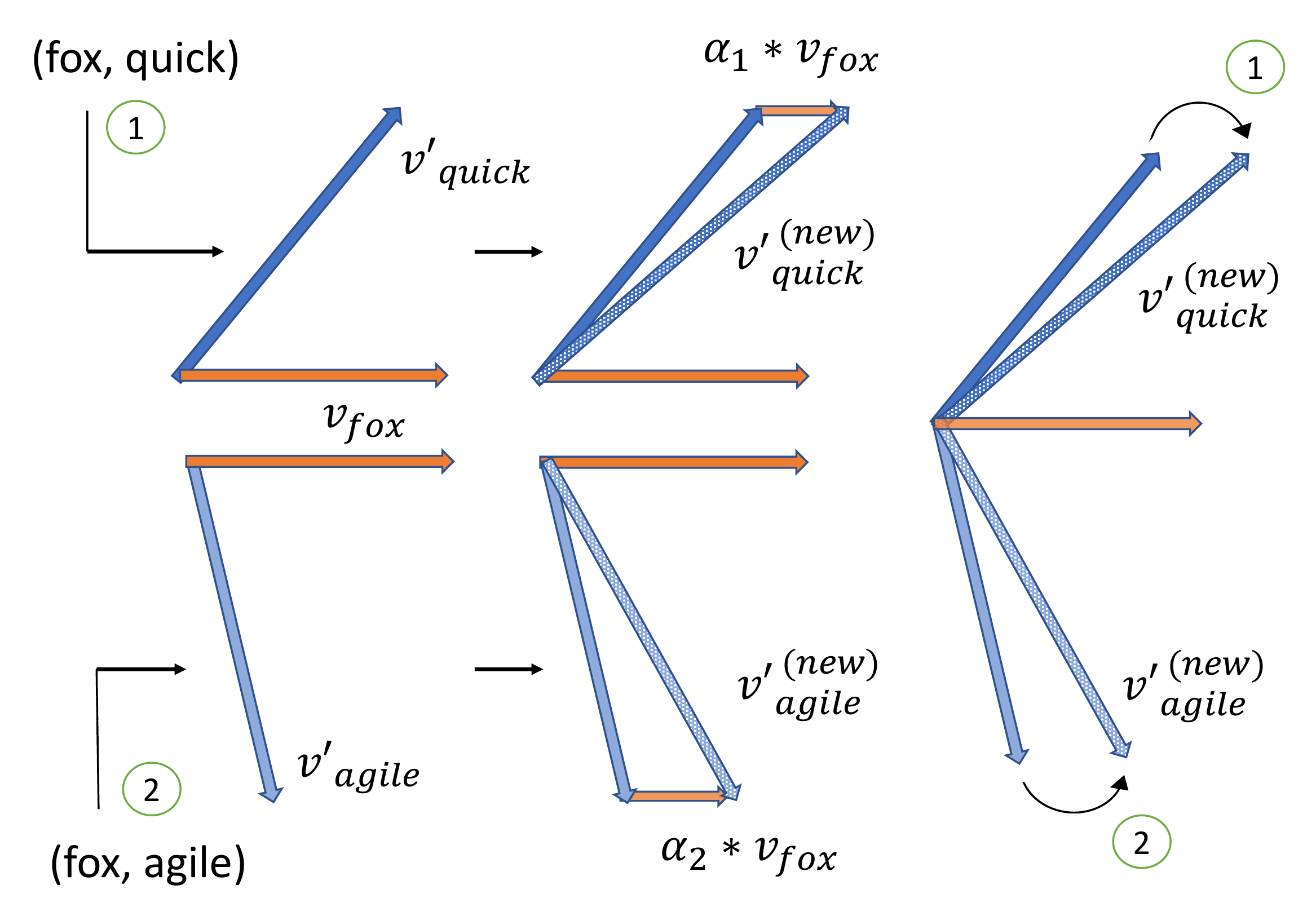}
\end{center}
\endminipage
\end{center}
\caption{\label{fig:scale} 
\emph{
Word2Vec samples a set window of neighbors around each word, introducing a fixed scale. Left:}  Word2Vec, a commonly used neural network for analysing language, categorises words in a window of fixed maximum size around a target word as its context words, thus introducing a \emph{set} scale. For each word, this generates several {\tt (target, context)} training samples \protect\cite<taken from>{mcco16u}. \textit{Right:} Word2Vec maintains an input and output vector representation for each word in its vocabulary, which are updated at each training sample. For example, when it sees the sample {\tt (fox,quick)} (labeled $1$), it brings the output vector for the context word {\tt quick} closer to the input vector for the target word {\tt fox}, and vice versa, which it would again do when it sees the sample {\tt (fox,agile)} (labeled $2$). However, by bringing the output vector for {\tt agile} closer to the input vector for {\tt fox}, it has brought the output vectors for {\tt agile} and {\tt quick} closer to each other, which both co-occur in the vicinity of the common word {\tt fox}.    }
\end{figure*}

A key aspect of Word2Vec is how the context around each target word is sampled, as this also introduces a definite scale into the algorithm. Word2Vec samples a window of words around the target word $w_{t}$, stretching out in both directions (shown in Fig \ref{fig:scale}). The size of the window is chosen randomly each for each new target word, but there is a maximal size $\beta$ which is usually defined as a fixed parameter before training commences. It can be shown that the resultant probability of choosing a neighboring word $w_{t\pm k}$ as a context word falls off linearly with the distance $k$ from the target, vanishing completely at $\beta$

\[p(w_{t\pm k}) = 1 - \frac{k - 1}{\beta}  \]

It is interesting to note that both the slope of this probability distribution and the reach of neighboring words accessible to it are governed completely by the choice of parameter $\beta$ - thus introducing a hard scale in the mechanics of the model. 

The vectors in the word2vec embeddings have also been seen to have some interesting features - vector arithmetic can often encode mappings of linguistic relations between the corresponding words. For example, vectors which transition from the vector representing the source word (eg. {\tt man}) to the destination word ({\tt woman}) for a particular relation, when then added to a different source word ({\tt king}), could take it very near to the intended destination word ({\tt queen}). This property of the Word2Vec embeddings could be used to test how well the embedding encodes different linguistic relationships, as explored in the next section.

\section{Methods}

\subsubsection{Corpus and Prepossessing} To train word2Vec, we used the {\tt enwik9} corpus from Matt Mahoney's repository\cite{matt06u}, containing preprocessed text from the first $10^9$ bytes of the Wikipedia dump dated Mar 3, 2006. Wikipedia was chosen to provide a rich representation of words which came from a diverse range of topics. The corpus consists of cleaned-up sentences which only retain text which would be visible to a human reader accessing a Wikipedia web page. The entirety of the text in the original article is retained, while converting all letters to lowercase. All numbers are converted to spelled out text (for e.g, $30$ becomes `three zero'). Hyperlinks are converted to retain only the description of the link accessible to the user, and letters which were not a-z were replaced by a single space. After preprocessing, the corpus contained 124 million words with a distinct vocabulary of 1.4 million words.  

\subsubsection{Training word2vec} We used the Continuous-Bag-of Words (CBOW) implementation of Word2Vec, written in C, from Mikolov's word2Vec Github repository \cite{miko17u}. word2vec utilizes a shallow three-layer neural network with one hidden layer. It maintains two active vector representations of each word in its vocabulary, called the `input' representation $v_i$ and the `outer' representation $v'_{i}$, encoded in the weight matrices between the layers. Both of these representations exist in the higher-dimensional vector space which would become the embedding. The hidden layer shares the same dimensionality, which we denote by $N$.

The CBOW algorithm tries to guess the target word given the set of context words surrounding that particular word. When the code is initialized, for each target word, word2vec generates (target, context) word pairs for each context word and passes each pair onto the neural network for training. Let us assume that, at a given time, the algorithm is given the pair $(w_O,w_I)$. Word2Vec starts with a one-hot representation $\mathbf{x}_{w_i}$, corresponding to the input context word $w_I$, as its input layer. A one-hot vector has dimension $V$ equaling the size of the vocabulary of the model, and only has a nonzero entry corresponding to the index of the word ($x_k = 1$ only when $k = I$, zero otherwise). 

The weight matrix $\mathbf{W}$ (dimension $V \times N$) projects from the input layer onto the hidden layer $\mathbf{h}$. This operation essentially generates the input vector representation of the input word $\mathbf{v}_{w_{I}}$

\[ 
\mathbf{h} = \mathbf{W}^{T} \mathbf{x}_{w_I} := \mathbf{v}^{T}_{w_{I}}
\]

The hidden layer then projects through another matrix, $\mathbf{W'}$ (dimension $N \times V$), generating a score $\mathbf{u}_{k}$ for each possible output word $w_k$

\[ \mathbf{u}_{k} = \mathbf{W'} \mathbf{h} = \mathbf{v'}_{w_{k}} \cdot \mathbf{v}_{w_{I}} \]

which effectively computes a dot product of the hidden layer with the output vector for each word $w_k$ in the vocabulary - representing how closely aligned each output vector $\mathbf{v'}_{w_k}$ is to the input vector $\mathbf{v}_{w_I}$. A softmax transformation finally converts this score into a posterior probability distribution. This becomes the corresponding entry $\mathbf{y}_k$ in the output layer of the network

\[ \mathbf{y}_{k} =  p(w_k|w_I) :=  \frac{\exp ({\mathbf{v'}_{k} \cdot \mathbf{v}_{I}})}{ \sum_{m = 1}^{V} \exp({ \mathbf{v'}_{m} \cdot \mathbf{v}_{I}})  } \]

This is word2vec's best guess about the chances of the word $w_k$ being the target word given that the word $w_I$ appeared in its context window. However, we had already started out with the actual answer $w_O$ for the target word. Through backpropagation, we now update the matrices $\mathbf{W}$ and $\mathbf{W'}$ (which generates the input and output representations respectively) so that the input vector for the context word ($\mathbf{v}_{w_I}$) and the output vector for the actual target word ($\mathbf{v'}_{w_O})$ moves closer to each other. All the output vectors not associated with the actual target word are moved further away from $\mathbf{v}_{w_I}$. At the end of the training, the space of input vectors $\mathbf{v}$ becomes the word embedding.

\subsubsection{Generating embeddings for different sampling scales}

The code sets up several threads of the Word2Vec process, which runs over the corpus simultaneously, training the weight matrices to minimize the loss function. The number of training iterations was increased to $30$ to improve consistency of similarity measurements across embeddings for each sampling scale. The parameters controlling for negative sampling and subsampling frequencies were left unchanged from the default values listed in the repository (refer to Table \ref{table:params}).

\begin{table}[H]
\begin{center}

\begin{tabular}{|c|c| }
\hline
Description of parameter & Value chosen \\
\hline
 Dimensionality of embedding & 200 \\ 
 Negative Sampling Loss (n) &  25 \\
 Subsampling frequency threshold & $10^{-4}$ \\
 Simultaneous threads running & 16 \\
 Number of training iterations & 30 \\
 \hline

\end{tabular}
\\[10pt]

 \caption{\label{table:params} Chosen values for different parameters used to implement the Continuous-Bag-of Words training in Word2Vec. }
    
\end{center}
\end{table}

\begin{figure*}[!ptb]

\minipage{0.99\linewidth}
    \includegraphics[width=0.999\columnwidth]{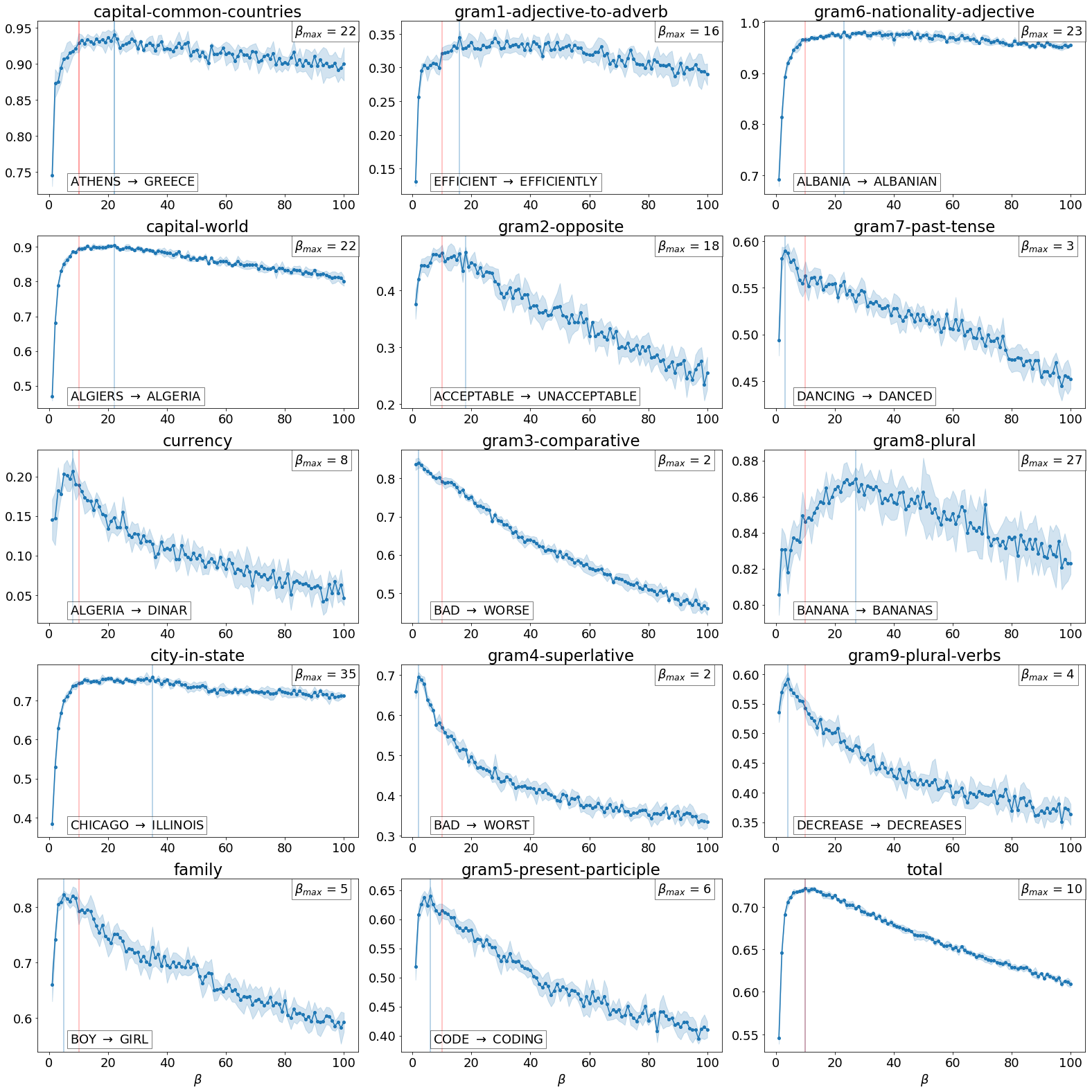}    % \begin{center}
    %  \scriptsize{From McCormick, C., 2016, Word2Vec Tutorial}   
    % \end{center}

\endminipage

\caption{\label{fig:acc} \emph{Different linguistic relations are encoded best at different sampling scales.} These graphs show Word2Vec's performance on the analogical reasoning task developed by Mikolov et al., for different linguistic relationships, as a function of the scale of context it is sampling ($\beta$). Each analogy uses 2 word pairs corresponding to a particular relationship. For example, a sample analogy in `capital-world' would ask, ``if {\tt France $\rightarrow$ Paris}, does India {\tt $\rightarrow$ Delhi}?", and the Word2Vec embedding is correct if by adding the direction vector for the first pair, $vec$({\tt France}) - $vec$({\tt Paris}), to the first word of the second pair, vec({\tt India}), we get a closest match to the second, i.e, $vec$({\tt Delhi}). The y-axis represents the fraction of correctly answered analogies, while the pale red line shows the average accuracy taken across tests.  Note the scale corresponding to maximum performance differs across panels, sometimes dramatically (shown in the upper-right corner in each panel and marked with the blue vertical line, while the position of the "best" scale taken across all tests is also marked in red).}

\end{figure*}

After setting up the algorithm, we generated different embeddings to capture different ranges of context for each word. To do this, we executed the training for values of scale parameters ranging from $\beta=1,2,3\dots 100$. Apart from the window size, all variables and chosen parameters were kept the same. For each scale, $10$ embeddings were generated to increase consistency and compute sampling statistics. The embeddings were analyzed by using the {\tt gensim} package \cite{RehuSojk10} in python.

\subsubsection{Encoding of linguistic relationships at different scales}

To analyze how the properties of the embedding changes with scale, we first tested how well the embeddings encoded different linguistic relationships as a function of the sampling scale used to generate it. We used Mikolov et al.'s analogical reasoning task to see if vector arithmetic can recognize linguistic maps between two words, for e.g {\tt boy} and {\tt girl}, and connect a different word through the same map, for e.g, {\tt son} to {\tt daughter}. This would be represented as the 4-tuple \{{\tt boy,girl,son,daughter}\} - for a general tuple of words $\{w_1,w_2,w_3,w_4\}$, the test amounts to checking if the vector closest to $v_{w_2} - v_{w_2} + v_{w_3}$, which is the direction vector going from $w_1$ to $w_2$ added to the vector for the word $w_3$, is closest to the vector for $w_4$. A list of such 4-tuples, analyzing maps from a total of $14$ different syntactic and semantic relations, was taken from the `questions-words.txt' in the repository and applied to the $30,000$ most frequent words found in the corpus. The fraction of correct choices for each linguistic relation and also the total fraction of correctly answered questions combining the performance across all of them was noted for each embeddings to gauge the variability of performance across different sampling scales.

\subsubsection{Capturing word neighborhoods at different scales}

We next examined scale-dependencies at a more local level by studying the neighborhood surrounding different word vectors. Words which are deemed more similar have higher cosine similarity in the vector space, and we studied how the cosine similarity of words surrounding a central word changed as the sampling scale of the embedding was varied. Efforts were made to distinguish systematic trends like the similarity of all neighbors shifting simultaneously, from more immediate changes affecting only a few neighbors, like certain neighbors becoming more similar to the central word, overtaking words higher on the list. Changes like the latter could be indicative of a change in the local semantic space, as we will see in the next section.

\subsubsection{Similarity statistics for neighbors at different scales}

Finally, we looked at distribution of cosine similarity among neighbors at different scales. Taking cue from the results of the previous analysis, we observe that different neighbors reach maximum similarity with the central word at very different scales. We aim to answer whether there is a preferred scale, or a clustering of scales, at which this happens - if this is the case, sampling the corpus around that scale of context would capture more information about a very high fraction of neighbors than sampling context at other scales. 

Solving this requires us to first decide on a set of neighbors for a particular central word, which we can track in embeddings from different sampling scales - with cosine similarities changing differently for different words, a word which qualifies as a neighbor at some sampling scale might be very far away at a different scale. We chose such a catalog by looking at the word vector of the central word in the embedding from each sampling scale, and record the $100$ closest neighbors to that word corresponding to that scale. The catalog of neighbors is then built by taking the collection of all the unique words in the combined record of these closest words generated from all sampling scales. Choosing the number of closest words chosen at each scale introduces a cutoff and could affect the distribution of scales corresponding to peak similarity - we study the effect of this by repeating our analysis with the catalog of neighbors chosen from $N = 5, 10, 20, 50, 100$ respectively. We then analyze the fraction of neighbors from our catalog which reach peak similarity with the central word at different scales.

\section{Results}

We are now ready to explore how the scale of sampling context around each word changes the structure of semantic space learned by Word2Vec. We first assess the performance of Word2Vec in encoding commonly encountered linguistic relationships using 4-vector analogical reasoning tasks \cite{MikoEtal13}, and examine if different relationships are best expressed at different scales. We then move from assessing the embeddings on a global level to looking at the individual neighborhoods of word vectors, and assess if the structure of the local semantic space itself is changing, or if the changes are purely systematic. Each neighbor is characterized by a sampling scale where it achieves maximum similarity with the central word. We then look at the distribution of sampling scales corresponding to peak similarity for neighborhoods of different words, and if there's a central scale around which they are clustered.

\begin{figure*} [!htb]
%\begin{center}

\minipage{0.47\linewidth}
\begin{center}
\includegraphics[width=0.95\textwidth]{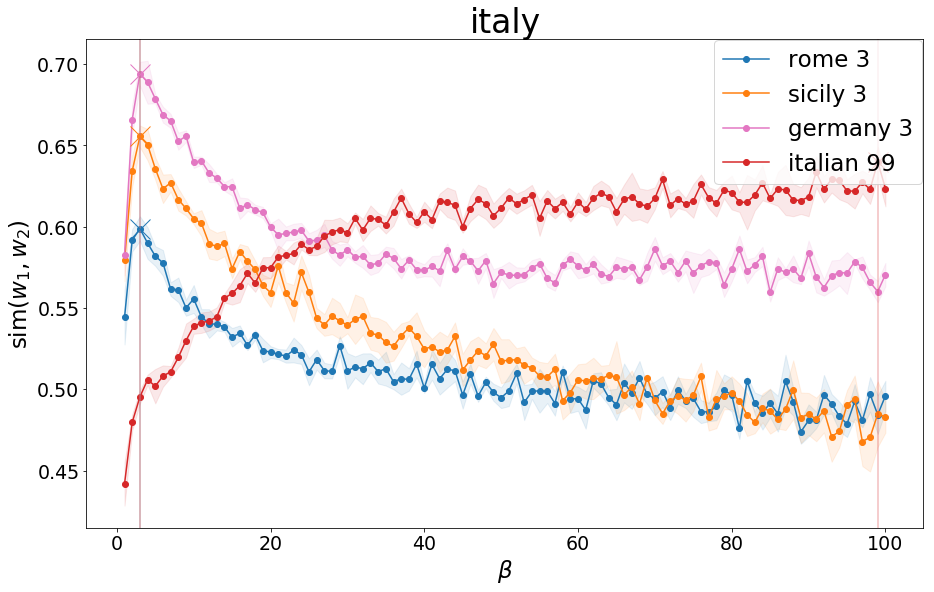}
\includegraphics[width=0.95\textwidth]{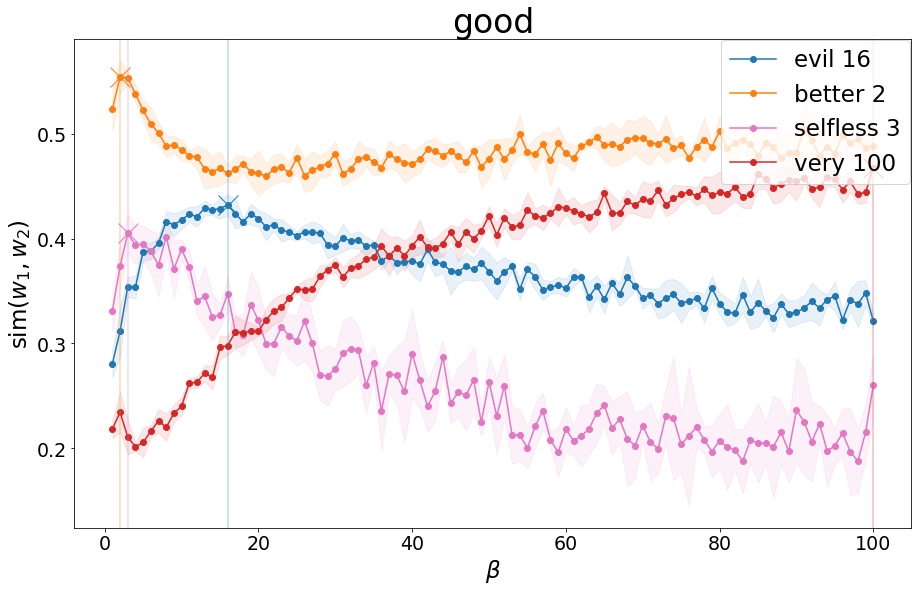}
\includegraphics[width=0.95\textwidth]{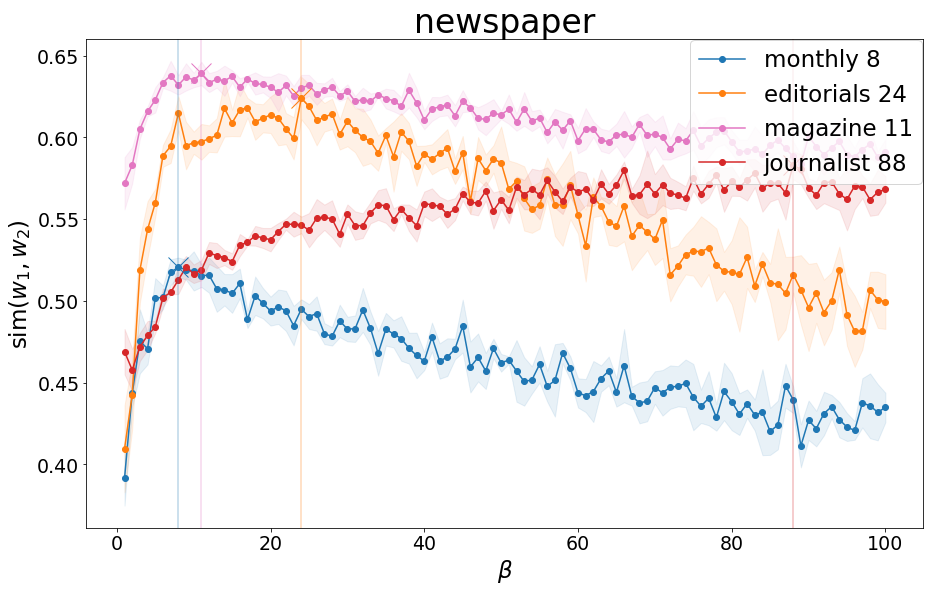}
\end{center}
\endminipage
\minipage{0.473\linewidth}
 \begin{center}

\includegraphics[width=0.95\textwidth]{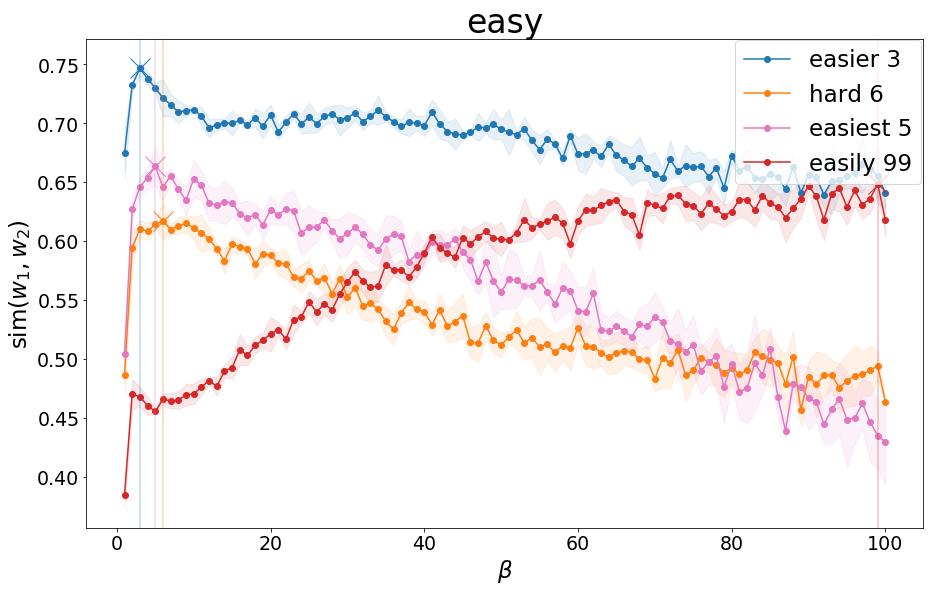}
\includegraphics[width=0.95\textwidth]{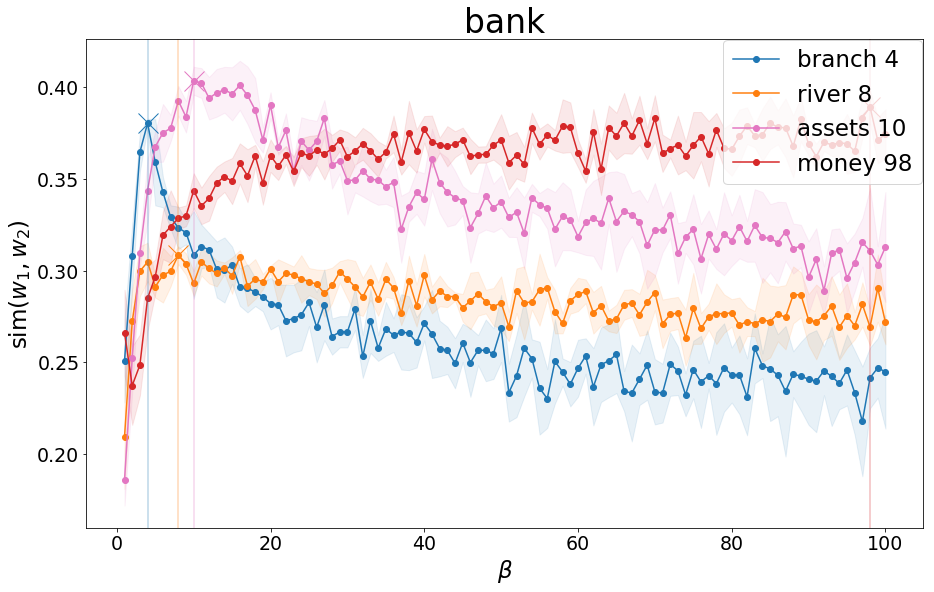}	
\includegraphics[width=0.95\textwidth]{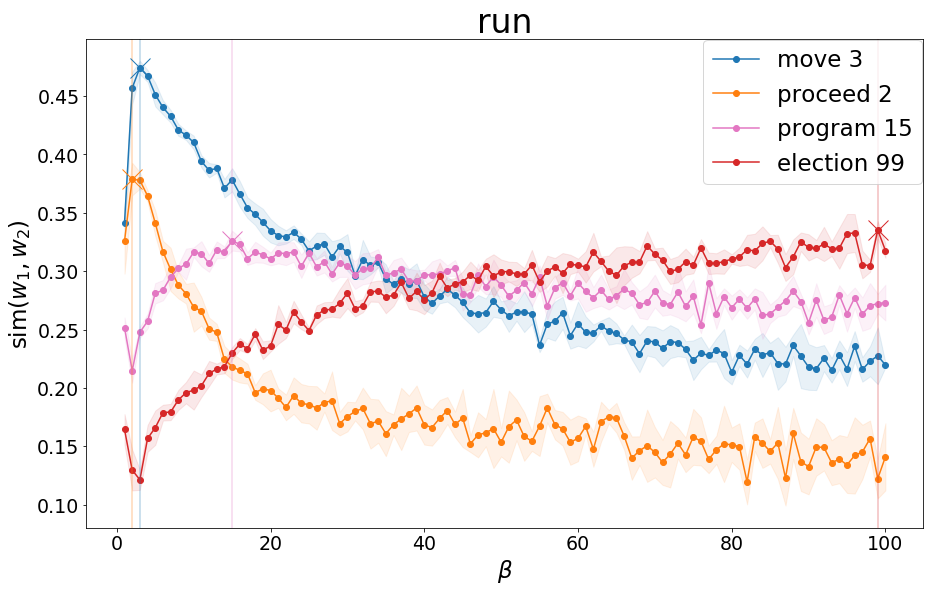}
\end{center}
\endminipage

%the shape of the semantic space depends on the scale that you choose
%if the space had the same shape, the ordering between neighbors would be preserved (modulo a scaling factor), if the space was changing, there would be some crossing over
% (the shaded regions show the inter-quartile variability taken over ten embeddings generated with the same $\beta$)
\caption{
% The neighbors of individual words change both qualitatively and quantitatively depending on the scale we look at
\emph{Relations between words change as a function of scale.} \textit{Row 1 and 2:} The plots show the cosine similarity of neighbors of a central word plotted as a function of the scale parameter($\beta$) of the embedding. The scale on the x-axis where each neighbor has the highest similarity is noted in the upper right corner beside the word. If the shape of the encoding vectors didn't change with $\beta$, the ordering between the neighbors should be preserved. We see here that the similarities for different words cross over at certain scales, implying that the shape of the semantic space depends on the scale that we choose. }
\label{fig:neighbors}
\end{figure*}

\subsection{Different Relationships at Different Scales}

Word2Vec has been shown to encode maps of different syntactic and semantic relationships \cite{MikoEtal13} - adding a direction vector going from an example of source to destination, $vec$({\tt France}) - $vec$({\tt Paris}), to another source word $vec$({\tt Germany}) can take us very close to the word vector for the correct destination $vec$({\tt Berlin}). In this section we explored whether the performance of encoding such relationships are modulated by the sampling scale of context chosen to generate the vector embedding. In Figure \ref{fig:acc}, we show the accuracy of the embeddings in answering the set of 4-vector analogical reasoning tasks for $14$ different linguistic relations, as developed to assess the performance of the model \cite{MikoEtal13}, but with the added exercise of testing them over embeddings generated using different sampling scales. The x-axis represents the sampling scale of the embedding being tested on, while the y-axis shows the fraction of the analogy tasks correctly answered by the embedding while testing a particular linguistic relation. 

It is surprising to note that the performance curves vary significantly among the different linguistic relations tested. There are qualitative differences in both the rise of the performance to the peak accuracy and the subsequent decay. There seem to a number of relations (e.g. `gram4-superlative', `currency', `family') for which peak accuracy is reached at fairly low scales and the performance seems to decay rapidly after achieving it. These are contrasted with some other relations (e.g. `gram1-adjective-to-adverb', `gram6-nationality-adjective', `city-in-state') which seem to reach maximum performance on the corresponding analogy tests slowly and at increasingly higher scales. There is a significant spread of scales where peak accuracy was achieved  – ranging from $\beta = 2$ for the `gram4-superlative' task to $\beta = 35$ for the `city-in-state' task, with quite a few clustered towards the higher end of the spectrum. This suggests that rather than any single scale, studying the embeddings over the entire spectrum of sampling scales might be required to capture different linguistic relationships.

\subsection{Different Neighborhoods at Different Scales}

In the last section, we found that different relationships are encoded in the Word2Vec embeddings at different scales -- however, that doesn't tell us if the local semantic space around the word vectors itself is changing. To answer this question, we looked at the neighbors of certain words and how the ordering of neighbors change as the size of the context sampled was varied, which is shown In Figure \ref{fig:neighbors}. The neighbors were chosen such that they reflect associates of the central word in different syntactic and semantic contexts. For e.g., the first two neighbors shown in the panel for the word {\tt run}  -- {\tt move} and {\tt proceed}, reflect the more common context of movement it is used in, while the latter ones exhibit the context of executing a program, and running as a candidate in an election, respectively. 

It is seen that the similarity curves of these neighbors with the central words shows marked qualitative and quantitative differences between them. Looking at the peak positions of the curves seems to indicate that neighbors achieve maximum similarity with the central word at very different sampling scales. There is often clustering of neighbors when they appear in similar contexts. The graph for the central word {\tt Italy} serves as a nice example of this, where {\tt Rome}, the capital of the country, is clubbed together with other related geographical locations {\tt Sicily} and {\tt Germany} ($\beta = 3$), while the adjective for being of the corresponding nationality {\tt Italian}, rises much more slowly to reach a maxima at $\beta = 99$. It would be very difficult to capture these intricate trends of behavior by sampling the text at any single, fixed scale.  

\begin{figure*}[!ptb]

\minipage{0.99\linewidth}
    \includegraphics[width=0.99\columnwidth]{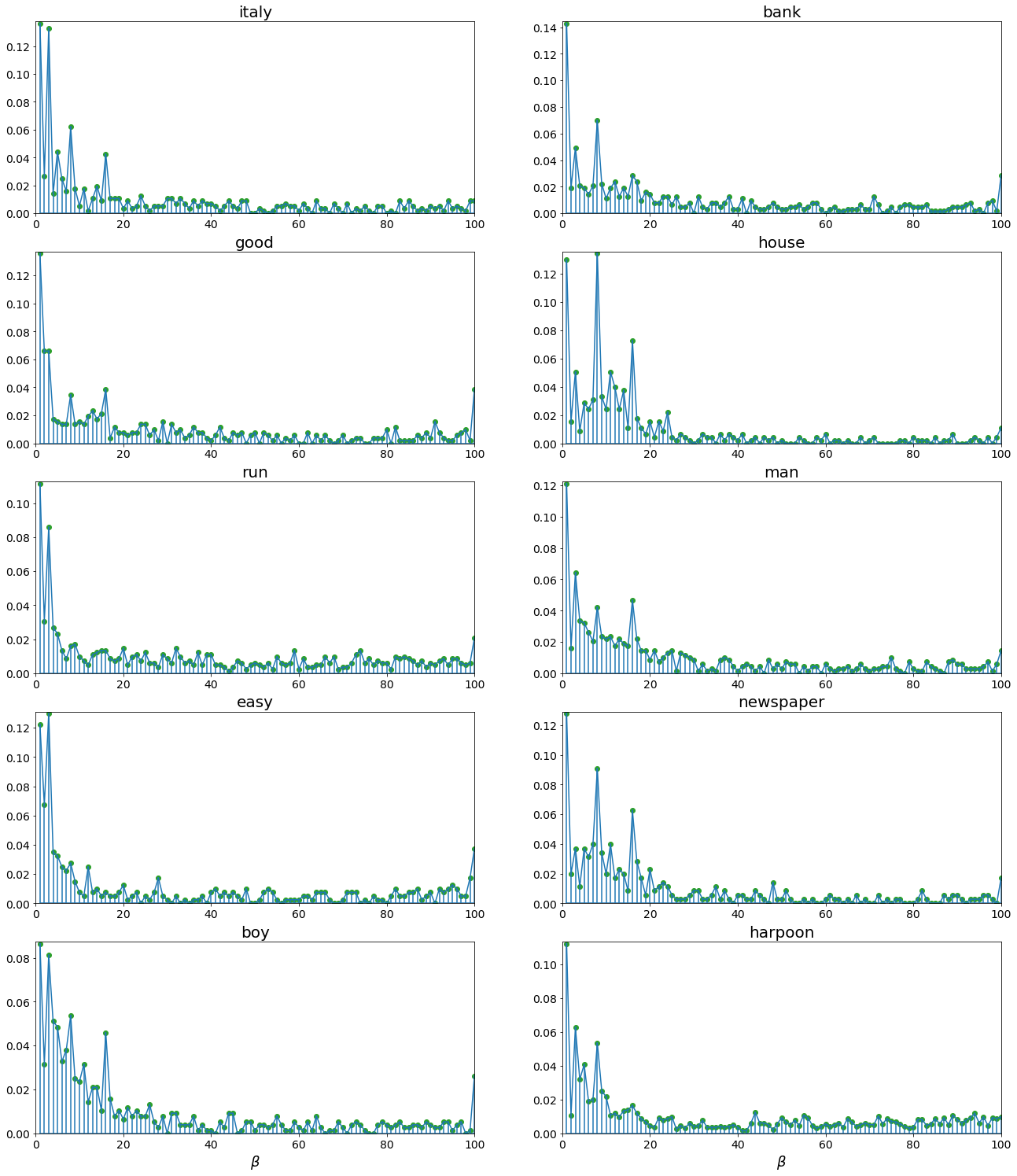}    % \begin{center}
    %  \scriptsize{From McCormick, C., 2016, Word2Vec Tutorial}   
    % \end{center}

\endminipage
% These graphs show the accuracy of retrieving 4-vector analogies on different syntactic and semantic relations as developed by Mikolov et al., plotted as a function of the scale parameter ($\beta$) of the embedding it is tested on.
\caption{\label{fig:hist} \emph{Neighbors of words show peak similarity at a wide range of scales.} These graphs show the normalized histograms of the distribution of sampling scales at which neighbors of a central word shows peak similarity with it. For example, {\tt finances}, {\tt money} and {\tt business} all reach peak similarity with the central word {\tt bank} at $\beta = 98$ and are thus all counted at that particular bin. The distribution of sampling scales is not centered around any one scale - the number of neighbors falls off slowly as the scale is varied. There seems to be a significant fraction of neighbors that reach peak similarity even at very high scales, which can also be seen from plotting the histograms on a log-log graph (inset in each figure). }
\end{figure*}

The differences between the curves also suggest a changing semantic space, as the sampling scale is varied. Changes in peak similarity alone in a systematic fashion doesn't imply a change in the the ordering of the neighbors - the similarity curves for successive neighbors could monotonically peak at higher scales and then decay in tandem, preserving the ordering between them. However, we also observe  lot of \emph{crossover} events, where the similarity curves of two different neighbors intersect at a particular sampling scale. This would imply that ordering of these neighbors would flip if we surveyed embeddings at scales going at lower to higher than the crossover scale. We see instances of this happening in each of the six panels, but the graph for the central word {\tt run} show a clear demonstration of this - the curves for the neighbors {\tt move}, {\tt program}, and {\tt election} all intersect at a very narrow region in $\beta$ and spread out again, essentially completely reversing the ordering of all three neighbors. These crossover events suggest that along with systematic changes, the shape of the semantic space around a word can itself drastically as the scale of sampling context is varied.

\subsection{Peak similarity for neighbors distributed at many different scales}

We see from the previous section that there are neighbors which reach maximum similarity with the central word at high sampling scales. It is not clear, however, whether such neighbors count for a very insignificant fraction of all the neighbors the word has at different scales, and therefore do not contribute much to the study of the semantic space around that word. To answer this question, we needed to study the distribution of scales at which peak similarity is reached for all neighbors of a word. Classifying whether a word as a neighbor requires us to choose a cutoff - we catalogued a list of the $100$ closest words to a central word at each sampling scale from $\beta = 1,2,3\dotsc 100$ and take the collection of all unique words gathered from all the scales. Changing the number of closest words gathered at each scale does not seem to qualitatively change the characteristics of the distribution for the variety of central words chosen except for the distribution becoming sharper as more neighbors are added to the pool.

In Figure \ref{fig:hist}, we plot the histograms of the scales at which maximum similarity with the central word was reached for each neighbor, plotted as the fraction of neighbors which have maximum similarity with the central word at the sampling scale denoted in the x-axis. The distribution for each central word has subtle differences, notably in the location and intensity where local maximas are observed, making each histogram a unique characterization of the word. More generally, however, none of the distributions seem to have a central scale, around which the peak sampling scales are clustered, refuting the notion that there is a single scale or a narrow range of preferred scales where most neighbors attain maximum similarity with the central words. Instead of an unimodal distribution, we see a gradual decline in the fraction of neighbors as the peak sampling scale is increased. There is a significant fraction of neighbors reaching peak similarity at the higher end of the scales we have studied, reminiscent of the trends seen in heavy-tailed distributions - which suggesting that capturing the full range of relationships between words would require studying the corpus at a spectrum of different scales.

\section{Discussion}

We have shown that the size of context while training Word2Vec can
significantly change the properties of the resultant embedding. It is seen
that to capture the semantic structure of different linguistic relationships,
context has to be captured at a wide spectrum of scales.   Because different
forms of information are carried at different scales,  the performance of
a language model depends on its sensitivity to scale.  
One can classify extant language models based on how they treat information at
different time scales.

\subsubsection{Language models with a single, fixed scale}

%There has been tremendous advances in natural language processing and language
%modeling over the last few years - however, a lot of widely used

Many contemporary language models sample context at fixed scales. For
instance, the introduction of self-attention mechanisms in the Transformer
architecture \cite{VaswEtal17} allowed it to look at the relationships between
words and model long-term dependencies without the need for recurrent units or
convolution. However, the algorithm trains on fixed-length segments of text,
and the self-attention looks at the contribution of all words within this
fragment to decipher the meaning of each word. This still constrains the
architecture to a fixed scale of context. It also introduces the problem of
context fragmentation \cite{DaiEtal19}, as the fragments scoop up a fixed length of symbols
without consideration of sentence structure or semantics. Thus the model
remains completely unaware of the context present in the previous segments
when it trains on the current segment, limiting its efficiency in looking at
the large-scale contexts present in the text. Transformers are used as
building blocks in many state-of the art language modeling architectures like
BERT \cite{DevlEtal18} from Google and GPT from OpenAI \cite{RadfEtAl18}.

The use of a fixed scale is seen also in older distributional models like latent semantic analysis (LSA) and the topic model \cite{GrifEtal07,LandDuma97}, which work with co-occurrence of words inside larger structures of text (documents). In LSA, the size of the document is chosen a priori (the default choice being $~300$ words), thus setting a fixed scale. The topic model is generative, as it tries to infer the distribution of words in each topic (a probability distribution over words) and distribution of topics in each document which would best account for the semantic structure in the source text. One still has to choose the number of topics beforehand, however, thus enforcing a scale. 

An effective scale is also seen in the syntagmatic-paradigmatic model \cite<SP,>{Denn04,Denn05}, which tries to extract structure from text by simultaneously keeping track of syntagmatic and paradigmatic associations between words. Syntagmatic associations are formed between words that occur together, like {\tt run} and {\tt fast}, as opposed to paradigmatic associations, which form between words which appear in similar context, like {\tt run} and {\tt walk}. The model keeps track of these by maintaining  memory traces which evaluates and stores different kinds of associations between words. However, these connections are computed between words within sentence-sized chunks, which sets a scale.

A fixed scale buffer has also carried over to moving window models like Word2Vec, and other vector embedding models like GloVe \cite<global vectors for word representation,>{PennEtal14}. Although the GloVe vectors are constructed to marry the best of both these worlds by calculating the co-occurrence matrix of a word around the context window of another word - but choosing the size of the context window still sets a scale.

\subsubsection{Language models that learn relevant timescales}

Other contemporary language models do not \emph{a priori} fix a scale, but
nonetheless have a set of scales that are learned \emph{via} training.
%Recurrent Neural Networks (RNNs) \cite{Elma90} have been one of the earliest
%and most widely known models to learn sequential dependencies in language
In recurrent neural networks \cite{Elma90,LawrEtal00,MikoEtal10,YaoEtal13}
the hidden state at a given time is computed as a function of
both the input at that step, and the hidden state for the preceding time step.
This allows the network to learn dependencies technically
without a fixed timescale \cite{AlpaEtal16}. It can be shown that a RNN learns
the relevant timescales it needs to maintain by updating the eigenvalues of
the weight matrix connecting the hidden states corresponding to sequential time steps.
However, that focusing on learning dependencies on only some preferred
timescales, even if not fixed, could lead a recurrent network to ignore
information at other timescales which could be essential in learning the
causal structure of the input data. 
Training RNNs to learn long-term
dependencies using standard gradient descent has also been shown to get
increasingly difficult as the time-scales to be captured become longer
\cite{BengEtal93,BengEtal94}. 

Long-Short Memory (LSTM) networks \cite{SchmHoch97} were introduced to tackle
both the vanishing and exploding gradient problem in RNNs \cite{HochEtal01}
and efficiently learn long-range dependencies \cite{HochSchm97}. They have
been successful in learning structure in language modeling
\cite{SundEtal12,WangJian15,SundEtal15} and has shown strong performance in
benchmarks \cite{Grav12,GersEtal99,GrefEtal16}. However, LSTMs can still
suffer from exploding gradients \cite{PascEtal12,LeZuid16,Gros17}. LSTMs have
also been shown to empirically use $200$ context words on average regardless
of the hyperparameters chosen, and start to disregard word order significantly
after the first 50 tokens \cite{khanEtal18}. More recent language modeling
architectures like Ulm-Fit \cite{HowaRude18} and contextualized word
representations like ELMo \cite{PeteEtal18} also use LSTM units as their
building blocks, implying that they could also suffer from a effective maximal
size of context.

\subsubsection{Towards scale-invariant language models}

We have seen that the statistical structure of language simultaneously carries different forms of information at different scales. However, many state-of-the-art
language models still address time scales as either a fixed buffer storing context, or attempt at learning relevant time scales as it parses through text. There has been recent efforts to combine features from both these classes \cite{DaiEtal19}, but the entire spectrum of time scales contained in the data are still not treated equivalently. 

Language models with fixed scale inherit this idea from short-term memory models from mid-twentieth century
psychology. George Miller's influential paper \cite{Mill56} argued the result that we can store "seven plus-or-minus two" simultaneous items of information in short term memory. The idea of short term memory as a fixed buffer store existing independently and separately from long term memory was further developed in the dual-store model \cite{AtkiShif68}. This classical view of short-term memory acting as fixed-capacity buffer in turn led to early computational models like HAL and BEAGLE \cite{JoneMewh07,LundBurg96} which featured a moving window which gathered context around a target word, a feature still used in many contemporary language models.

In the intervening decades, ideas in psychology and neuroscience have evolved towards a scale-invariant
working memory \cite{BalsGall09,ChatBrow08,Gibb77}.
Biological neural networks exhibit a wide of time scales and carry information
about many different scales, including systematic changes at the scale of
seconds, minutes, hours and even days.
\cite{BernEtal11,MauEtal18,RubiEtal15,CaiEtal16,BrigEtal19}. Neuronal ensembles has been seen to fire at increasing latencies following a stimulus with a gradually increasing firing spread \cite{PastEtal08,Eich14,SalzEtal16}. These \emph{time cells} behave like a short term memory, retaining information not only about the timing but also the identity of the stimulus \cite{TigaEtal18a,CruzEtal18},  but have a spectrum of time scales. It is possible to build cognitive models from scale-invariant time cells that describe behavior, underscoring the usefulness of a scale-invariant representation of temporal history in models of cognition \cite{HowaEtal15}.
%J Cog Neuro?

How would one incorporate these insights into a new generation of language models?
It seems like a new
generation of language models employing scale-free buffers \cite{ShanHowa13},
which can store information from exponentially long timescales at the cost of
discounting temporal accuracy,  and might be able to learn structure simultaneously from different scales of context. Such a model would not have to direct attention only to a fixed subset of scales, either predetermined or learned, but would be able to attend equally to the entire spectrum of observed timescales, extracting useful predictive information about scale-dependent relationships in natural language.

% Spectrum of scales.  Each treated equivalently.

\section{Conclusion}

 In this work, we have investigated how the scale of sampling context around each word changes the structure of semantic space learned by Word2Vec. It is seen that different relationships can have markedly different performances at different scales and they seem to be best encoded at a large spectrum of sampling scales. Looking at the individual neighborhoods of word vectors, we find that the local semantic space around words seems to change qualitatively and that the ordering of neighbors around a word can be drastically different based on the scale that context is sampled. We also find that a significant fraction of neighbors for a given central words reach maximal similarity with it even at high sampling scales. The statistics of such maximal scales does not seem to be peaked at any central scale but rather seem to follow a slowly decaying distribution as the sampling scales are increased. These results seem to indicate that there is not a preferred scale to study language - there is different information about the structure of the semantic space at different scales, which would be better analyzed by scale-invariant models of statistical learning.

\section{Acknowledgments}

We gratefully acknowledge discussions with Kim Stachenfeld, Per Sederberg and Brandon Jacques. This work was supported by IIS 1631460 and Google AIFRA.

\bibliographystyle{apacite}
\bibliography{bibdesk}% Produces the bibliography via BibTeX.

\end{document}